\documentclass[pdflatex,sn-vancouver-num]{sn-jnl}

\usepackage{graphicx}%
\usepackage{multirow}%
\usepackage{amsmath,amssymb,amsfonts}%
\usepackage{amsthm}%
\usepackage{mathrsfs}%
\usepackage[title]{appendix}%
\usepackage{xcolor}%
\usepackage{textcomp}%
\usepackage{manyfoot}%
\usepackage{booktabs}%
\usepackage{algorithm}%
\usepackage{algorithmicx}%
\usepackage{algpseudocode}%
\usepackage{listings}%
\usepackage{tabularx}
\usepackage{float}
\usepackage{placeins}
\begin{document}

\title{CiteGuard-RAG: A Validation-Centered AI System for Evidence-Grounded Question Answering}

\author*{\fnm{Sumit} \sur{Barua}}\email{sumit.barua@wmich.edu}
\author{\fnm{Guan} \sur{Hong}}
\author{\fnm{Halil} \sur{Dursunoglu}}
\author{\fnm{Charles} \sur{Rodgers}}
\author{\fnm{Alvis} \sur{Fong}}

\affil {\orgdiv{Department of Computer Science}, \orgname{Western Michigan University}, \orgaddress{\city{Kalamazoo}, \state{MI}, \postcode{49008}, \country{USA}}}

% -----------------------------
% Abstract
% -----------------------------

\abstract{
Retrieval-augmented generation (RAG) can improve access to complex information; however, retrieving evidence alone does not ensure that answers are grounded, citation-valid, or appropriately refused. This paper introduces \textit{CiteGuard-RAG}, a validation-centered AI system for evidence-grounded question answering. The system integrates hybrid semantic--lexical retrieval, citation-constrained generation, sentence-level grounding validation, and single-pass regeneration. Validation is used at runtime to determine whether a candidate answer should be accepted, refused, or regenerated before final delivery.

CiteGuard-RAG is evaluated on 400 questions across a controlled housing-law dataset, PrivacyQA, and CUAD. In the controlled evaluation, it achieves 99.1\% retrieval accuracy, 98.3\% grounded-answer accuracy, and 98.3\% citation validity, with no validation-detected hallucinations. Ablation results show that grounded-answer accuracy drops sharply when validation is removed, even when retrieval accuracy remains unchanged. External evaluation shows that while citation validity remains strong, evidence utilization, span alignment, and refusal calibration become harder under domain shift.

These findings indicate that trustworthy RAG systems require explicit validation between retrieval and final answer delivery. CiteGuard-RAG provides a practical architecture for linking retrieval, generation, citation checking, abstention, and regeneration in high-stakes information access.
}


\maketitle

% -----------------------------
% Introduction
% -----------------------------
\section{Introduction}
\subsection{Background and Motivation}
Large language models (LLMs) have improved natural language understanding and generation, but their use in high-stakes information access remains limited by hallucination, weak source traceability, and unreliable refusal behavior \citep{ziwei2023, zihuai2024}. These limitations are especially important in legal question answering, where answers must remain faithful to source documents and unsupported claims can mislead users.

RAG addresses part of this problem by conditioning LLM outputs on retrieved evidence rather than relying only on parametric memory \citep{fan2024survey,ram2023incontext}. However, retrieval success does not guarantee grounded generation. Even when relevant passages are retrieved, a model may omit key evidence, overgeneralize from partial context, merge retrieved text with parametric knowledge, or produce unsupported elaborations \citep{mallen2023trust}.

Legal information access further amplifies these challenges because legal corpora are large, heterogeneous, and structurally complex \citep{niklaus2024multilegalpile}. Legal texts often contain hierarchical organization, specialized language, exceptions, and argumentation patterns that complicate retrieval, chunking, and evidence grounding \citep{habernal2024mining}. As a result, trustworthy legal RAG systems require explicit mechanisms for verifying whether generated claims are supported by cited evidence, rather than assuming that retrieval alone ensures reliability \citep{hindi2025legalrag}.

\subsection{Literature Review and Research Gaps}
\subsubsection{Retrieval Strategies}
Hybrid retrieval methods that combine lexical methods such as BM25 with dense semantic retrieval have been widely studied in open--domain question answering, and the findings show that lexical methods are better at exact term matching, while dense methods are better at capturing semantic similarity but can retrieve topically related but non-specific evidence \citep{arivazhagan2023hybrid, chirkova2024multilingualrag}. In the legal domain, transformer--based rerankers designed for very long queries and documents showed considerable performance gains while keeping computational efficiency \citep{askari2024rprs}. Retrieval partitioning strategies have also proven that the organization and incorporation of recovered content greatly improve the quality of downstream generations \citep{wang2024mrag}. Despite these developments, the interaction of retrieval strategies and grounding quality remains insufficiently understood in legal RAG systems.

\subsubsection{Grounding and Hallucination}
A recurrent difficulty for RAG systems is that retrieval quality alone does not guarantee grounded generation; models can hallucinate when retrieved evidence is marginally relevant or contradicts parametric knowledge \citep{mallen2023trust}. This difficulty is exacerbated by the “lost-in-the-middle” effect, where LLMs suffer substantial degradation when crucial information is located in the middle of long input contexts. This phenomenon suggests that models do not robustly employ all returned evidence \citep{liu2024lost}. To mitigate hallucinations in legal QA, constrained RAG techniques have been developed that leverage aspect-based constraints throughout the retrieval and generation stages, demonstrating improvements in terms of accuracy and relevance of retrieved documents \citep{nguyen2024consrag}. Likewise, systems based on natural language inference (NLI) and reinforcement learning have demonstrated the capacity of entailment-based validation to enhance the factual consistency of generated text \citep{roit2023factually}. However, prior work does not have a unified framework to impose grounding during generation and verify it at fine--grained, sentence--level resolution simultaneously.

\subsubsection{Evaluation}
The evaluation of RAG systems is difficult due to the connection between retrieval and generation. Standard end-to-end metrics often conflate retrieval mistakes with generation failures, such that system weaknesses are obscured \citep{salemi2024evaluating, zhang2024llmeval}. Benchmarking studies have also demonstrated that RAG systems still suffer from robustness, information integration, and negative rejection behavior \citep{chen2024benchmarking}. Recent surveys in the legal domain indicate the importance of specialized evaluation metrics that measure faithfulness, citation validity, and hallucination at fine granularity \citep{hindi2025legalrag}. In-context retrieval-augmented models show enhanced grounding and source attribution \citep{ram2023incontext}, but there is no systematic evaluation of grounding reliability across system components.

\subsubsection{Research Gaps}
The existing literature leaves four main gaps. First, it remains unclear how effectively RAG systems can enforce evidence-grounded and citation-valid answers in high-stakes information access tasks, especially under domain shift. Second, the effect of lexical, semantic, and hybrid retrieval strategies on downstream grounding and hallucination behavior is not yet well characterized. Third, existing validation approaches often remain post-hoc, answer-level, or evaluation-oriented, leaving limited understanding of how sentence-level citation and evidence checks can control whether a candidate answer is accepted, regenerated, or refused before final delivery. Fourth, retrieval, generation, validation, and regeneration are often evaluated as separate components, even though their interaction determines final answer reliability.

\subsection{Proposed Approach}
\begin{figure}[H]
\centering
\includegraphics[width=\textwidth]{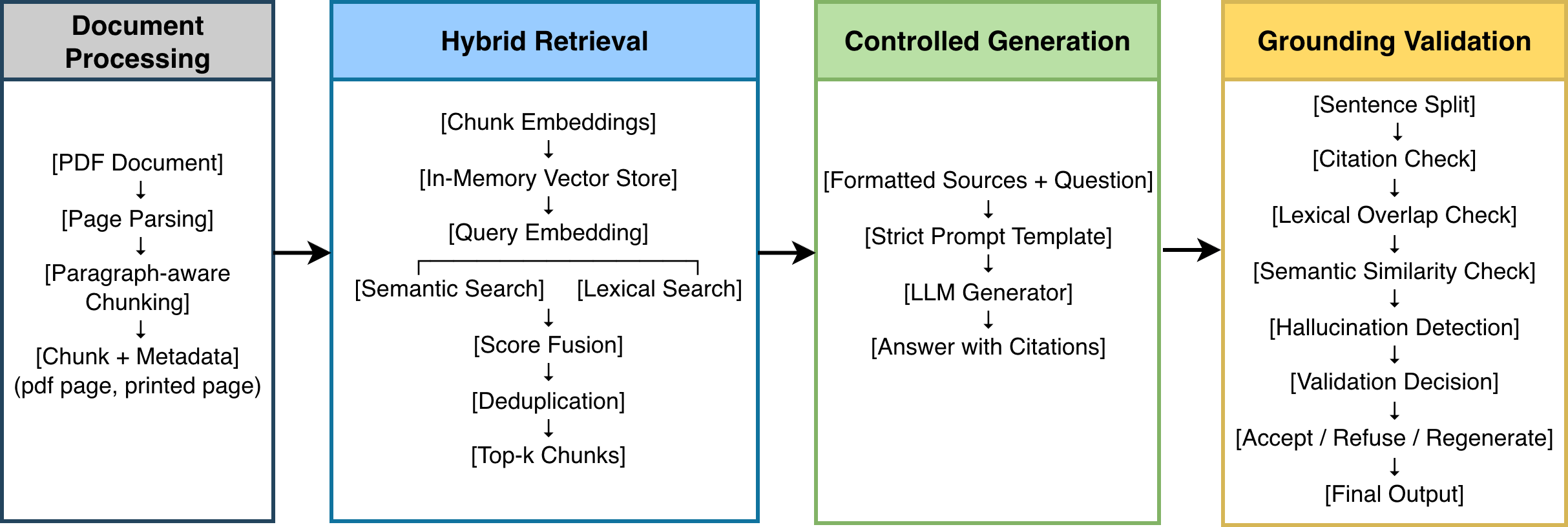}
\caption{CiteGuard-RAG pipeline for evidence-grounded question answering. The framework integrates paragraph-aware chunking, hybrid semantic--lexical retrieval with score fusion, citation-constrained generation, post-generation grounding validation, and one-pass validation-guided regeneration.}
\label{fig}
\end{figure}

In this paper, we propose \textit{CiteGuard-RAG}, a validation-centered intelligent information system for evidence-grounded question answering. The framework combines page-aware document parsing, paragraph-aware chunking, hybrid semantic--lexical retrieval, citation-constrained generation, sentence-level grounding validation, and one-pass validation-guided regeneration. The overall system design is shown in Figure~\ref{fig}.

The central idea is to treat grounding as a runtime control problem rather than only as a post-hoc evaluation step. The pipeline first generates an internal candidate answer and then checks whether its claims are supported by cited retrieved evidence before any response is delivered. The validation result determines whether the candidate is accepted as the final answer, revised through one-pass regeneration, or replaced with a standardized refusal.

\subsection{Contributions}
Motivated by the research gaps above, this study examines how retrieval, generation, validation, refusal, and regeneration interact in an evidence-grounded RAG system. The evaluation focuses on whether citation-grounding validation improves answer reliability, how retrieval and design choices affect grounding behavior, and what failure modes emerge under controlled and external domain-shift settings. The main contributions of this work are:
\begin{enumerate}
    \item \textit{CiteGuard-RAG}, a validation-centered intelligent information system that connects retrieval, citation-constrained generation, grounding validation, abstention, and one-pass regeneration for evidence-grounded question answering.

    \item A sentence-level citation-grounding validation mechanism that jointly checks citation validity, lexical evidence overlap, and semantic support before final answer delivery.

    \item Empirical analysis of retrieval--generation--validation interaction, showing that high retrieval accuracy alone does not guarantee grounded or citation-valid answers.

    \item Multi-setting evaluation over 400 questions across a controlled housing-law dataset, PrivacyQA, and CUAD, allowing in-domain performance to be compared with external domain and task-shift behavior.

    \item Failure case analysis identifying key reliability bottlenecks, including false refusal, citation/evidence-compliance failures, answer-span mismatch, and evidence-utilization errors.
\end{enumerate}

% -----------------------------
% Methodology
% -----------------------------
\section{Methodology}
\subsection{System Overview}
Given a document corpus \(D\) and a user query \(q\), the system retrieves a ranked evidence set \(E \subset D\) and generates an internal candidate answer \(a_c\). The candidate answer is not delivered directly. Instead, each substantive claim is checked against the retrieved evidence and its associated citation. The final output is either a validated answer \(a\) or a standardized refusal. The pipeline is defined as:
\[
\begin{aligned}
D &\rightarrow \text{Parsing} \rightarrow \text{Chunking} \rightarrow \text{Indexing} \\
  &\rightarrow \text{Retrieval} \rightarrow \text{Candidate Generation} \rightarrow \text{Validation} \\
  &\rightarrow \text{Accept/Regenerate/Refuse}.
\end{aligned}
\]

Algorithm~\ref{alg:citation_constrained_rag} summarizes the proposed citation-constrained RAG workflow, including retrieval, evidence filtering, validation, and one-pass regeneration.

\begin{algorithm}[!htbp]
\caption{CiteGuard-RAG With Runtime Validation and One-Pass Regeneration}
\label{alg:citation_constrained_rag}
\begin{algorithmic}[1]
\Require Query \(q\), document corpus \(D\), retriever \(R\), generator \(G\), validator \(V\)
\Ensure Validated final answer \(a\) or standardized refusal

\State Parse \(D\) into page-aware text units
\State Construct paragraph-aware evidence chunks \(\{C_i\}\)
\State Retrieve hybrid-ranked evidence candidates \(E' = R_{\mathrm{hybrid}}(q,\{C_i\})\)
\State Deduplicate candidates and select top-\(k\) evidence \(E \subseteq E'\)

\If{\(E\) is insufficient}
    \State \Return standardized refusal
\EndIf

\State Generate internal candidate answer \(a_0 = G(q,E)\) using citation-constrained prompting
\State Compute validation result \(v_0 = V(a_0,E)\)

\If{\(v_0\) passes}
    \State \Return validated final answer \(a_0\)
\Else
    \State Generate regenerated candidate \(a_1 = G(q,E,v_0)\) using validation feedback
    \State Compute validation result \(v_1 = V(a_1,E)\)
    \If{\(v_1\) passes}
        \State \Return validated final answer \(a_1\)
    \Else
        \State \Return standardized refusal
    \EndIf
\EndIf

\end{algorithmic}
\end{algorithm}

\subsection{Document Parsing and Chunk Construction}
Input legal documents are processed using page-aware text extraction to preserve evidence provenance. Each extracted page is represented as:
\[
P_i = \{\text{text}_i, \text{pdf\_page}_i, \text{printed\_page}_i\},
\]
where \(\text{text}_i\) is normalized page text, \(\text{pdf\_page}_i\) is the internal PDF page index, and \(\text{printed\_page}_i\) is an optional human-visible page number recovered from the document when available. When printed page numbers cannot be reliably detected, the system falls back to the canonical PDF page index. This preserves traceability across retrieval, citation display, and evaluation.

Documents are segmented using a paragraph-aware chunking strategy. Paragraph boundaries are preferred when preserved by PDF extraction; otherwise, a sentence-level fallback is applied using punctuation-based boundary detection. Adjacent text units are aggregated into chunks under a maximum token budget with overlap to preserve local context across chunk boundaries. To avoid semantically diffuse evidence, chunk expansion is constrained by page span, and low-information fragments such as navigation text, tables of contents, and very short segments are removed. Each final chunk is stored as:
\[
C_j = \{\text{text}_j, \text{chunk\_id}_j, \text{pdf\_range}_j, \text{printed\_range}_j\}.
\]

\subsection{Hybrid Retrieval}
The retrieval layer combines dense semantic retrieval and sparse lexical retrieval. Dense retrieval encodes each chunk \(C_i\) and query \(q\) using a sentence-transformer embedding model \(f(\cdot)\):
\[
\mathbf{c}_i = f(C_i), \qquad \mathbf{q} = f(q).
\]
Semantic relevance is computed using cosine similarity:
\[
\mathrm{Sim}_{\mathrm{semantic}}(q,C_i)
=
\frac{\mathbf{q}\cdot\mathbf{c}_i}{\|\mathbf{q}\|\|\mathbf{c}_i\|}.
\]
In parallel, lexical relevance is estimated using BM25:
\[
\mathrm{Score}_{\mathrm{lexical}}(q,C_i)
=
\sum_{t \in q}
\mathrm{IDF}(t)
\cdot
\frac{tf_{t,i}(k_1+1)}
{tf_{t,i}+k_1\left(1-b+b\cdot\frac{|C_i|}{\mathrm{avgdl}}\right)}.
\]
Semantic and lexical scores are min--max normalized and fused as:
\[
\mathrm{Score}_{\mathrm{hybrid}}
=
\alpha \hat{s}_{\mathrm{semantic}}
+
(1-\alpha)\hat{s}_{\mathrm{lexical}},
\]
where \(\alpha\) controls the semantic--lexical trade-off. This hybrid formulation supports exact legal phrase matching while retaining semantic flexibility for paraphrased queries.

\subsection{Candidate Filtering and Optional Reranking}
Retrieved candidates are deduplicated using normalized text similarity to reduce repeated evidence caused by overlapping chunks and improve context diversity before generation.

A lightweight heuristic reranker based on query overlap and legal cue words was evaluated as an optional module. However, because reranking reduced performance in the controlled evaluation, the final system uses deduplicated hybrid retrieval without reranking. Neural cross-encoder rerankers are not used, as this study focuses on lightweight, locally deployable RAG components.

\subsection{Controlled Evidence-Grounded Generation}
Retrieved chunks are formatted as numbered source blocks and passed to the language model using a constrained prompt. The prompt requires the model to answer only from retrieved sources, cite supporting evidence using inline source markers, avoid external legal knowledge, and refuse when sufficient evidence is unavailable. The standardized refusal response is:
\begin{quote}
\emph{The document does not contain sufficient information to answer this question.}
\end{quote}

This prompting strategy constrains the language model to function as a source-grounded answer generator rather than a free-form legal advisor. For external datasets with different document types, the same principle is preserved: answers must be grounded in retrieved evidence, citations must point to valid retrieved sources, and unsupported claims trigger refusal or regeneration.

\subsection{Post-Generation Validation and Regeneration}
\begin{figure}[!htbp]
\centering
\includegraphics[width=0.80\linewidth, keepaspectratio]{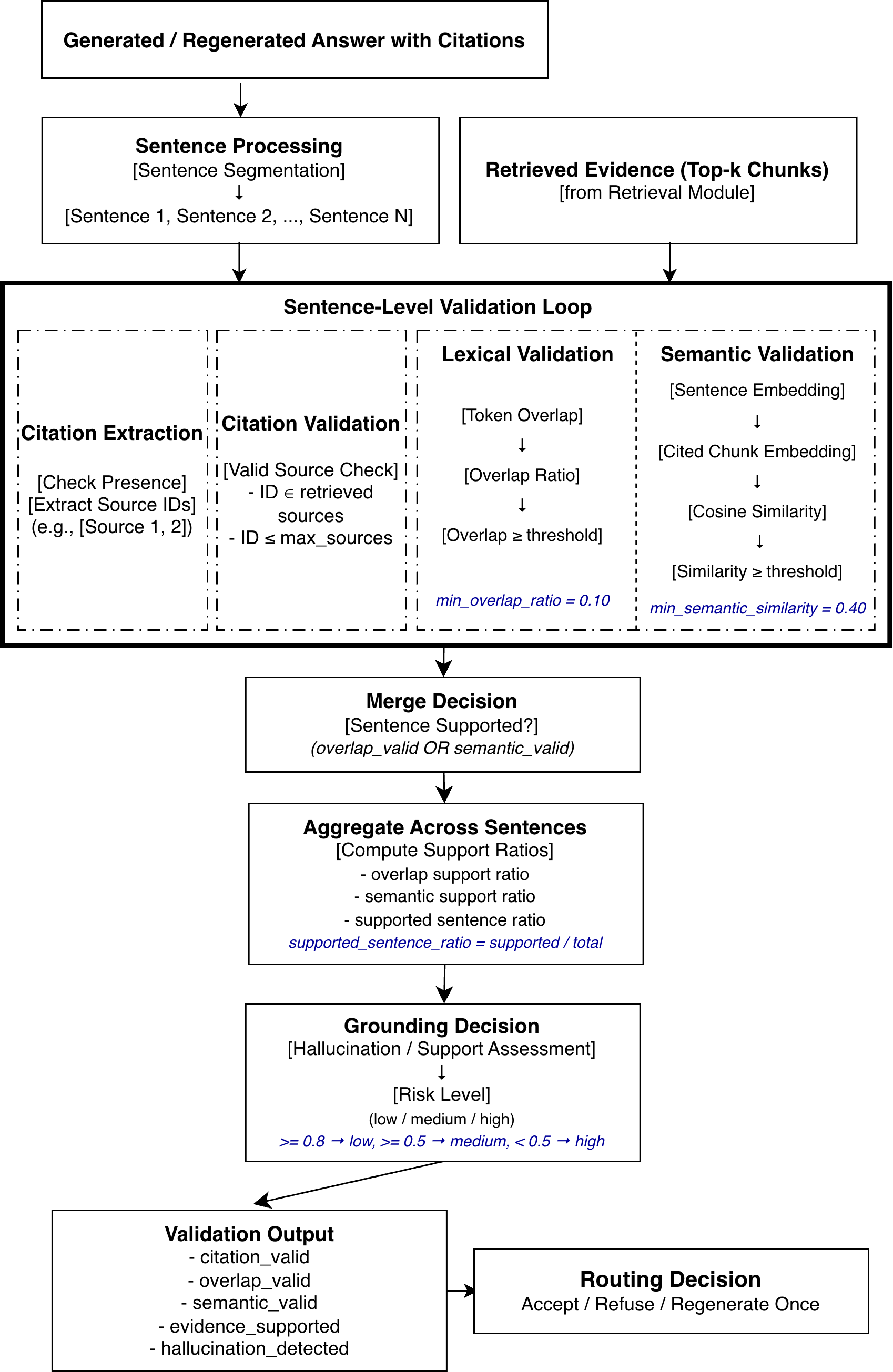}
\caption{Sentence-level grounding validation pipeline. The workflow verifies citations, lexical overlap, and semantic similarity against retrieved evidence, then routes the candidate output toward acceptance, one-pass regeneration, or refusal.}
\label{fig:validation}
\end{figure}

The validation layer checks whether each internally generated candidate answer is traceable to retrieved evidence before any answer is delivered as the final response. Candidate answers, including regenerated candidates, are decomposed into sentences, and each sentence is evaluated for citation validity and grounding support. Figure~\ref{fig:validation} illustrates the sentence-level validation workflow used to route candidate outputs toward final acceptance, one-pass regeneration, or refusal.

A citation is considered valid only if it refers to one of the retrieved source blocks. Lexical grounding is computed using token overlap between a candidate sentence and its cited evidence:
\[
\mathrm{Overlap}(s,e)=\frac{|T_s \cap T_e|}{|T_s|},
\]
where \(T_s\) and \(T_e\) denote normalized token sets for the candidate sentence and cited evidence, respectively.

Semantic grounding is computed using sentence embeddings:
\[
\mathrm{Sim}_{\mathrm{semantic}}(s,e)
=
\frac{\mathbf{s}\cdot\mathbf{e}}{\|\mathbf{s}\|\|\mathbf{e}\|}.
\]
A sentence is considered supported if it satisfies citation validity and passes either the lexical or semantic grounding threshold. The lexical-overlap and semantic-similarity thresholds were set to 0.10 and 0.40, respectively, based on pilot validation runs to balance strict hallucination control against valid paraphrased answers. These values are tunable implementation parameters rather than fixed theoretical constants.

A candidate response is flagged as hallucinated if it contains one or more unsupported claims. The supported sentence ratio is computed as:
\[
\mathrm{SupportedRatio}=\frac{n-u}{n},
\]
where \(n\) is the total number of candidate sentences and \(u\) is the number of unsupported sentences.

When validation detects a grounding-risk signal, the system performs a single regeneration attempt using a stricter evidence-grounded prompt. Regeneration is triggered by invalid citation behavior, weak evidence support, unsupported content, malformed citation behavior, or refusal despite strong retrieved context. The regenerated candidate is accepted only if it improves validation compliance or, when evidence is insufficient, produces a clean refusal. Regeneration is limited to one pass to control inference cost and avoid unbounded correction loops. Thus, validation affects both safety and coverage: unsupported candidate answers are blocked before delivery, but valid answers may also be rejected when the system behaves conservatively.

\subsection{Controlled Evaluation Set}
The controlled evaluation set consists of 150 question--answer pairs derived from 18 housing-law documents across multiple U.S. states and cities. Each question is associated with gold evidence pages and a refusal label indicating whether the document contains sufficient information to answer the query. The dataset includes 123 answerable questions and 27 unanswerable questions. This design supports evaluation of both answer generation and abstention behavior.

To reduce formulation bias, the evaluation set combines 100 human-authored questions with 50 LLM-assisted questions that were manually reviewed, edited, and verified against the source documents. Questions are stratified across factual, procedural, reasoning, and edge-case categories. Table~\ref{tab:question_types} reports the question-type distribution.

\begin{table}[!htbp]
\centering
\caption{Distribution of controlled evaluation question types}
\label{tab:question_types}
\small
\begin{tabular}{l c p{8.0cm}}
\hline
\textbf{Category} & \textbf{Count} & \textbf{Description} \\
\hline
Factual & 51 & Direct questions asking for explicit information stated in the source document. \\
Procedural & 40 & Questions asking about steps, rights, obligations, deadlines, or required actions. \\
Reasoning & 32 & Questions requiring interpretation or combination of related evidence from the retrieved text. \\
Edge Cases & 27 & Questions involving exceptions, unavailable evidence, ambiguous conditions, or refusal behavior. \\
\hline
\end{tabular}
\end{table}

\subsection{External Validation Design}
External validation is conducted using two complementary datasets. PrivacyQA \citep{ravichander} is used as the primary external QA-style validation dataset because it preserves the natural document-grounded question-answering structure of the controlled housing-law evaluation while shifting to privacy-policy documents. PrivacyQA consists of user questions about privacy policies with expert-labeled relevant policy segments. We evaluate a fixed 200-question subset and treat expert-labeled relevant segments as gold evidence, allowing assessment of retrieval coverage, citation validity, evidence support, and regeneration beyond the housing-law corpus.

CUAD \citep{hendrycks2021cuad} is used as a supplementary contract-domain stress test rather than as a direct natural-QA benchmark. CUAD differs from the controlled evaluation because it asks for contract clause identification and span-like evidence, introducing both legal domain shift and task-formulation shift. We evaluate a fixed 50-question stratified subset containing answerable and unanswerable clause-extraction examples to assess retrieval coverage, citation validity, abstention behavior, hallucination control, and regeneration under contract-domain shift.

\subsection{Implementation Details}
The system was implemented as a modular Python pipeline. Dense embeddings were generated using BAAI/bge-base-en-v1.5 \cite{Xiao2023}, while lexical retrieval used BM25 \cite{robertson2009probabilistic}. Hybrid retrieval combined min--max normalized semantic and lexical scores using a 0.65/0.35 weighting scheme. The final controlled configuration used 600-token chunks with 120-token overlap, top-\(k=8\), and up to four retrieved chunks passed to the generator. The heuristic reranker was disabled in the final system because it reduced controlled-setting performance.

Generation was performed using locally deployable instruction--tuned models under deterministic decoding with temperature set to 0. When validation failed, the system allowed at most one regeneration attempt using validation feedback. For external validation, the same RAG architecture, citation requirements, validation framework, and one-pass regeneration mechanism were retained, with dataset-specific preprocessing applied only to align PrivacyQA and CUAD with their native evidence structures.

\subsection{Evaluation Protocol}
Each controlled housing-law test instance is represented as:
\[
(q, y_{\mathrm{refuse}}, G),
\]
where \(q\) is the query, \(y_{\mathrm{refuse}}\in\{0,1\}\) indicates whether the system should abstain, and \(G\) denotes the gold evidence pages. For PrivacyQA, \(G\) corresponds to expert-labeled relevant policy segments. For CUAD, \(G\) corresponds to annotated contract evidence spans or relevant clause text.

Evaluation separates retrieval coverage, answer correctness, citation grounding, hallucination control, abstention behavior, and regeneration behavior. This separation allows retrieval failures to be distinguished from generation, validation, and refusal-calibration failures. The main metric groups are summarized in Table~\ref{tab:evaluation_metrics}.

Two hallucination-related metrics are reported. \textit{Hallucination Rate} measures broader answer-level failures, including incorrect, overextended, benchmark-inconsistent, or evidence-inconsistent final responses under the dataset-specific evaluation criteria. \textit{Unsupported Hallucination Rate} is narrower and measures final responses that contain claims not supported by the retrieved and cited evidence. Thus, a response may count toward the broader hallucination rate when it fails a benchmark answer criterion, while not counting as an unsupported hallucination if its claims remain grounded in cited evidence.

\begin{table}[!htbp]
\centering
\caption{Evaluation metric groups}
\label{tab:evaluation_metrics}
\small
\begin{tabular}{p{3.2cm} p{8.8cm}}
\hline
\textbf{Metric group} & \textbf{Metrics and purpose} \\
\hline

Retrieval coverage & Page Hit@$k$, Span/Segment Hit@$k$, and Context Recall@$k$ measure whether gold evidence is retrieved and included in the generation context. \\

Answer correctness & Grounded Answer Accuracy and Token F1 measure whether the final answer is correct, evidence-aligned, and sufficiently overlaps with gold evidence or expert-labeled segments. \\

Citation and evidence support & Citation Validity and Evidence Support Rate measure whether final answers cite valid retrieved sources and satisfy lexical or semantic grounding checks. \\

Hallucination control & Hallucination Rate measures broader incorrect, overextended, benchmark-inconsistent, or evidence-inconsistent final responses. Unsupported Hallucination Rate measures claims not supported by retrieved and cited evidence. \\

Abstention behavior & True Refusal Rate, False Refusal Rate, and False Answer Rate measure whether the system appropriately refuses when evidence is insufficient and avoids answering unanswerable questions. \\

Regeneration behavior & Regeneration Attempt Rate and Regeneration Accepted Rate measure how often validation triggers regeneration and how often the regenerated candidate is accepted as the final output. \\

\hline
\end{tabular}
\end{table}

% -----------------------------
% Results
% -----------------------------
\section{Results}
\subsection{Overall System Performance}
Table~\ref{tab:overall_performance} reports the performance of the final selected system configuration on the controlled housing-law evaluation set.

\begin{table}[!htbp]
\centering
\caption{Overall system performance under the final selected configuration}
\label{tab:overall_performance}
\begin{tabular}{l c}
\hline
\textbf{Metric} & \textbf{Value} \\
\hline
Retrieval Accuracy & 99.1\% \\
Grounded Answer Accuracy & 98.3\% \\
Citation Validity & 98.3\% \\
Evidence Support Rate & 98.3\% \\
Hallucination Rate & 0.0\% \\
False Refusal Rate & 1.8\% \\
True Refusal Rate & 96.4\% \\
Regeneration Rate & 0.9\% \\
\hline
\end{tabular}
\end{table}

The system achieves high retrieval accuracy, grounded answer accuracy, citation validity, and evidence support. No hallucination is detected under the implemented validation criteria, while the false-refusal rate remains low. The low regeneration rate indicates that most candidate answers satisfy the grounding constraints on the first generation pass.

To assess whether this behavior depends on a specific generator, we also evaluated Mistral \cite{jiang_mistral_2023} and Qwen \cite{Bai2023} under the same retrieval, prompting, and validation configuration. As shown in Table~\ref{tab:llm_benchmark}, both models achieve identical performance across the primary reliability metrics.

\begin{table}[!htbp]
\centering
\caption{Controlled-setting comparison of generation models}
\label{tab:llm_benchmark}
\begin{tabular}{l c c c c}
\hline
\textbf{Model} & \textbf{Grounded Acc.} & \textbf{Citation Validity} & \textbf{Hallucination} & \textbf{True Refusal} \\
\hline
Mistral & 98.3\% & 98.3\% & 0.0\% & 96.4\% \\
Qwen & 98.3\% & 98.3\% & 0.0\% & 96.4\% \\
\hline
\end{tabular}
\end{table}

The identical model-level results suggest that, in the controlled setting, system behavior is primarily determined by retrieval, prompting, and validation constraints rather than by generator-specific variability.

\subsection{Component and Configuration Analysis}
Table~\ref{tab:embeddings} compares three embedding models under the same hybrid retrieval configuration. BAAI/bge-base-en-v1.5 achieves the strongest overall performance, with the highest retrieval accuracy and grounded answer accuracy and the lowest hallucination rate.

\begin{table}[!htbp]
\centering
\caption{Embedding model comparison}
\label{tab:embeddings}
\begin{tabular}{l c c c c}
\hline
\textbf{Embedding Model} & \textbf{Retrieval} & \textbf{Grounded} & \textbf{Hallucination} & \textbf{False Refusal} \\
\hline
all-MiniLM-L6-v2 \cite{Wang2020} & 90.8\% & 89.2\% & 3.1\% & 6.2\% \\
intfloat/e5-base-v2 \cite{Wang2022} & 90.8\% & 90.8\% & 3.1\% & \textbf{0.0\%} \\
BAAI/bge-base-en-v1.5 \cite{Xiao2023} & \textbf{99.1\%} & \textbf{98.3\%} & \textbf{0.0\%} & 1.8\% \\
\hline
\end{tabular}
\end{table}

Table~\ref{tab:retrieval_modes} compares lexical-only, semantic-only, and hybrid retrieval using the selected BGE embedding model. Hybrid retrieval outperforms both single-mode baselines, suggesting that legal QA benefits from combining exact phrase matching with semantic similarity.

\begin{table}[!htbp]
\centering
\caption{Comparison of retrieval strategies}
\label{tab:retrieval_modes}
\begin{tabular}{l c c c c}
\hline
\textbf{Mode} & \textbf{Retrieval Acc.} & \textbf{Grounded Acc.} & \textbf{Hallucination} & \textbf{Exact Hit} \\
\hline
Lexical Only & 84.6\% & 83.1\% & 4.6\% & 81.5\% \\
Semantic Only & 92.3\% & 90.8\% & 3.1\% & 84.6\% \\
Hybrid (0.65/0.35) & \textbf{99.1\%} & \textbf{98.3\%} & \textbf{0.0\%} & \textbf{99.1\%} \\
\hline
\end{tabular}
\end{table}

The validation layer has the largest effect on generation reliability. As shown in Table~\ref{tab:validation}, disabling validation leaves retrieval accuracy unchanged but substantially reduces grounded answer accuracy and citation validity while increasing hallucination.

\begin{table}[!htbp]
\centering
\caption{Effect of validation layer}
\label{tab:validation}
\begin{tabular}{l c c c c}
\hline
\textbf{Setting} & \textbf{Retrieval Acc.} & \textbf{Grounded Acc.} & \textbf{Citation Validity} & \textbf{Hallucination} \\
\hline
Validation ON & 99.1\% & \textbf{98.3\%} & \textbf{98.3\%} & \textbf{0.0\%} \\
Validation OFF & 99.1\% & 8.7\% & 12.3\% & 18.6\% \\
\hline
\end{tabular}
\end{table}

This result is important for legal QA because citation validity and evidence support cannot be inferred from retrieval success alone; they require explicit post-generation verification.

Reranking was evaluated as an optional retrieval-stage refinement. However, as shown in Table~\ref{tab:reranker}, the heuristic reranker reduced both retrieval accuracy and grounded answer accuracy while increasing false refusals. The final configuration therefore disables reranking.

\begin{table}[!htbp]
\centering
\caption{Effect of reranker}
\label{tab:reranker}
\begin{tabular}{l c c c c}
\hline
\textbf{Setting} & \textbf{Retrieval Acc.} & \textbf{Grounded Acc.} & \textbf{False Refusal} & \textbf{Exact Hit} \\
\hline
Reranker OFF & \textbf{99.1\%} & \textbf{98.3\%} & \textbf{1.8\%} & \textbf{99.1\%} \\
Reranker ON & 87.7\% & 86.2\% & 3.1\% & 83.1\% \\
\hline
\end{tabular}
\end{table}

Chunk size and hybrid weighting also affect system behavior. Table~\ref{tab:chunking} shows that a 600-token chunk size provides the best trade-off between context completeness and retrieval specificity.

\begin{table}[!htbp]
\centering
\caption{Effect of chunk size}
\label{tab:chunking}
\begin{tabular}{l c c c c}
\hline
\textbf{Chunk Size} & \textbf{Retrieval Acc.} & \textbf{Grounded Acc.} & \textbf{False Refusal} & \textbf{Hallucination} \\
\hline
400 & 93.8\% & 92.3\% & \textbf{1.5\%} & 1.5\% \\
600 & \textbf{99.1\%} & \textbf{98.3\%} & 1.8\% & \textbf{0.0\%} \\
800 & 95.4\% & 93.8\% & 3.1\% & 1.5\% \\
\hline
\end{tabular}
\end{table}

Table~\ref{tab:weights} evaluates different semantic--lexical score combinations within hybrid retrieval. The 0.65/0.35 configuration achieves the best overall balance, producing the highest retrieval accuracy and grounded answer accuracy while eliminating detected hallucination under the validation criteria. More semantic--heavy retrieval at 0.80/0.20 performs strongly but increases hallucination risk, indicating that maximizing semantic dominance does not necessarily yield the safest legal-QA configuration.

\begin{table}[!htbp]
\centering
\caption{Effect of hybrid semantic/lexical weighting}
\label{tab:weights}
\begin{tabular}{c c c c c c}
\hline
\textbf{Semantic} & \textbf{Lexical} & \textbf{Retrieval Acc.} & \textbf{Grounded Acc.} & \textbf{Hallucination} & \textbf{False Refusal} \\
\hline
0.80 & 0.20 & 96.9\% & 96.9\% & 4.6\% & \textbf{0.0\%} \\
0.65 & 0.35 & \textbf{99.1\%} & \textbf{98.3\%} & \textbf{0.0\%} & 1.8\% \\
0.50 & 0.50 & 90.8\% & 90.8\% & 1.5\% & 3.1\% \\
0.35 & 0.65 & 89.2\% & 89.2\% & 1.5\% & 1.5\% \\
\hline
\end{tabular}
\end{table}

\subsection{External Validation and Stress-Test}
Table~\ref{tab:external_results} reports external validation results on PrivacyQA and CUAD. These datasets evaluate whether the framework maintains retrieval, citation, grounding, refusal, and regeneration behavior outside the controlled housing-law setting.

\begin{table}[!htbp]
\centering
\caption{External validation and stress-test results}
\label{tab:external_results}
\begin{tabular}{l c c}
\hline
\textbf{Metric} & \textbf{PrivacyQA} & \textbf{CUAD Stress Test} \\
\hline
Evaluation Size & 200 questions & 50 questions \\
Retrieval Recall@$k$ & 93.37\% & 100.00\% \\
Citation Validity & 100.00\% & 100.00\% \\
Evidence Support Rate & 81.33\% & 36.00\% \\
Answer Accuracy & 68.50\% & 54.00\% \\
True Refusal Rate & 76.47\% & 92.00\% \\
False Refusal Rate & 18.67\% & 56.00\% \\
False Answer Rate & 23.53\% & 8.00\% \\
Hallucination Rate & 22.00\% & 26.00\% \\
Unsupported Hallucination Rate & 0.00\% & 6.00\% \\
Regeneration Attempt Rate & 30.00\% & 28.00\% \\
Regeneration Accepted Rate & 29.50\% & 16.00\% \\
\hline
\end{tabular}
\end{table}

On PrivacyQA, the system achieves high retrieval recall, citation validity, and evidence support, with no unsupported hallucination detected under the validation criteria. Regeneration is triggered for nearly one-third of the questions, indicating that runtime validation remains active under external evaluation.

The CUAD stress test shows a larger shift in evidence-support and refusal behavior. Retrieval recall and citation validity remain high, but evidence support decreases and false refusal increases. Regeneration is also less often accepted as the final output.

\subsection{Failure Case Analysis}
To examine residual errors, we analyze representative failures across the controlled evaluation and external validation. In the controlled setting, residual errors are rare and primarily involve false refusal or citation/evidence-compliance failures rather than unsupported generation. Under external validation, errors become more frequent and dataset-specific: PrivacyQA errors include gold-segment mismatch and false-answer behavior, while CUAD errors are dominated by over-refusal and clause-span extraction difficulty.

Observed residual errors are grouped into five categories: (i) false refusal, where the system abstains despite available evidence; (ii) citation or evidence-compliance failure, where an otherwise supported answer fails strict citation or validation requirements; (iii) segment mismatch, where a concise supported answer does not fully match the benchmark's longer gold evidence segment; (iv) false answer, where the system answers despite insufficient evidence; and (v) extraction or span-selection failure, where relevant evidence is retrieved but not used completely or precisely.

Table~\ref{tab:failure_cases} provides representative examples of these residual error types.

\begin{table}[!htbp]
\centering
\caption{Representative residual error cases across controlled and external settings}
\label{tab:failure_cases}
\begin{tabular}{l l p{3.5cm} p{3.5cm}}
\hline
\textbf{Setting} & \textbf{Type} & \textbf{Observed Behavior} & \textbf{Cause} \\
\hline

Controlled & False refusal & Evidence is available but the system abstains & Conservative refusal behavior under validation \\

Controlled & Citation compliance & Supported answer fails strict citation or validation formatting & Strict citation-control requirements \\

PrivacyQA & Segment mismatch & Supported concise answer does not fully match gold segment & Gold labels are long policy segments rather than canonical answers \\

PrivacyQA & False answer & System answers when evidence is insufficient for the query & Difficulty distinguishing partial evidence from sufficient evidence \\

CUAD & Over-refusal & Relevant clause is retrieved but system abstains & Clause-extraction task differs from natural QA \\

CUAD & Extraction failure & Relevant clause is retrieved but answer is incomplete & Clause spans are dispersed across context \\

\hline
\end{tabular}
\end{table}

% -----------------------------
% Discussion
% -----------------------------
\section{Discussion}
\label{sec:discussion}

\subsection{Findings and Interpretation}
The results show that trustworthy evidence-grounded RAG requires more than strong retrieval. Reliable answer generation depends on the interaction among retrieval, citation-constrained generation, post-generation validation, and refusal behavior. In the controlled housing-law evaluation, CiteGuard-RAG retrieves relevant evidence and constrains generation to cited, source-supported content under the implemented validation criteria.

The generator comparison further suggests that controlled-setting performance is not driven mainly by a specific language model. Mistral and Qwen produce effectively identical results under the same retrieval, prompting, and validation configuration, indicating that the retrieval and validation layers strongly constrain model behavior. This is important for intelligent information systems where reproducibility, deployment control, and privacy-sensitive local execution may be required.

External validation provides a more realistic view of system behavior under task and domain shift. PrivacyQA shows that citation validity and evidence support can remain strong when the framework is transferred to another document-grounded QA setting, although strict segment-overlap scoring reduces measured answer accuracy. CUAD presents a harder contract-domain stress test, where retrieval and citation behavior remain strong but false refusal and evidence-support errors increase. Overall, the framework generalizes most strongly in retrieval coverage, citation control, and unsupported-hallucination prevention, while answer-span selection and abstention calibration remain more sensitive to domain and task formulation.

\subsection{Retrieval and Representation}
The retrieval results show that hybrid semantic--lexical retrieval provides the most reliable evidence access. Lexical retrieval helps preserve exact legal and policy wording, while semantic retrieval improves matching when questions paraphrase the source text. The best performance is achieved when these signals are balanced rather than used independently.

The weighting and embedding ablations further show that retrieval design affects downstream grounding, not only evidence recall. Excessive semantic weighting increases hallucination risk, while stronger representations improve both retrieval and answer reliability. This indicates that evidence representation is a core reliability factor in RAG-based intelligent information systems.

\subsection{Validation, Grounding, and Regeneration}
The validation layer is the main mechanism for controlling unsupported generation. When validation is removed, retrieval accuracy remains unchanged, but grounded answer accuracy and citation validity decrease sharply. This confirms that retrieval alone is not sufficient for trustworthy evidence-grounded QA.

The controlled 0.0\% hallucination rate should be interpreted under the study's operational definition: no unsupported claims were detected relative to retrieved and cited evidence. This does not mean that all possible legal or factual errors were eliminated. The remaining controlled-setting errors were mainly false refusals and citation or evidence-compliance failures rather than unsupported final answers.

The identical Mistral and Qwen results suggest that, in the controlled setting, system behavior is primarily shaped by retrieval, prompting, and validation constraints rather than by generator-specific variability. Regeneration further supports reliability by giving the system one opportunity to revise candidate answers that fail validation, although external results show that it cannot fully solve over-refusal or span-selection errors.

\subsection{System Design Trade-offs}
The ablation results show that reliability depends on practical design choices. Moderate chunking provides the best balance between preserving enough context and avoiding irrelevant evidence. Smaller chunks can fragment legal rules, while larger chunks can add noise.

The reranking experiment also shows that added complexity does not always improve reliability. The heuristic reranker suppressed some relevant answer-bearing chunks, so the final system uses deduplicated hybrid retrieval without reranking. Overall, the results support optimizing the full retrieval--generation--validation pipeline rather than isolated components.

\subsection{External Validation Insights}
The external evaluations clarify the type of generalization achieved by CiteGuard-RAG. PrivacyQA suggests that the pipeline transfers reasonably well to another document-grounded legal/policy QA setting, particularly in citation control and unsupported-hallucination prevention. This supports the role of runtime validation beyond the controlled housing-law corpus.

CUAD reveals a different limitation. Unlike the controlled housing-law and PrivacyQA settings, CUAD is closer to contract clause extraction than natural question answering. It therefore introduces both domain shift and task-formulation shift. Although the system retrieves relevant evidence and maintains valid citations, it becomes more conservative and produces more false refusals. This suggests that clause-level contract extraction requires more precise span-selection mechanisms than the current answer-generation pipeline provides. CUAD should therefore be interpreted as a stress test of robustness under task shift rather than as a direct comparison to the controlled QA setting.

Together, the two external evaluations show that retrieval coverage and citation validity generalize more robustly than answer extraction and refusal calibration. This distinction is important for evidence-grounded RAG systems: a system may retrieve the right evidence and cite valid sources, yet still fail to produce the exact answer expected by a benchmark. Reliable high-stakes question answering therefore requires both evidence access and evidence utilization.

\subsection{Limitations and Future Work}
Several limitations should be acknowledged. First, the controlled evaluation uses a curated housing-law corpus, and the external evaluations are limited to PrivacyQA and CUAD. Although these datasets provide meaningful evidence under domain and task shift, broader testing across statutes, regulations, case law, administrative guidance, and legal-aid documents would strengthen generalizability claims.

Second, the external benchmarks differ in answer format. PrivacyQA provides expert-labeled policy segments rather than canonical short-form answers, while CUAD focuses on clause-level contract evidence. As a result, answer accuracy and token-overlap scores should be interpreted alongside citation validity, evidence support, false-answer behavior, and unsupported hallucination.

Third, the validation framework relies on threshold-based lexical and semantic similarity checks. These checks help detect unsupported outputs, but they may miss implicit entailment, complex legal reasoning, or nuanced paraphrases. Future work should explore stronger validation methods based on natural language inference, learned evidence verification, and answer-span selection before generation.

Fourth, regeneration is limited to a single attempt for efficiency and reproducibility. Bounded multi-pass regeneration may improve recovery from difficult generation failures, but it would require careful stopping criteria to avoid excessive inference cost or unstable outputs.

Finally, the evaluated generators are general-purpose instruction-tuned models without legal-domain adaptation. Future work should compare domain-specialized and stronger general-purpose models under the same citation, grounding, refusal, and regeneration constraints.

% -----------------------------
% Conclusion
% -----------------------------
\section{Conclusion}
This work presented \textit{CiteGuard-RAG}, a validation-centered intelligent information system for evidence-grounded question answering combining hybrid semantic--lexical retrieval, citation-constrained generation, sentence-level grounding validation, abstention, and one-pass validation-guided regeneration. Results from the controlled housing-law evaluation show that strong retrieval alone is not sufficient for reliable answer generation; explicit post-generation validation is needed to enforce citation validity, evidence support, and hallucination control. External evaluation on PrivacyQA and CUAD further shows that retrieval coverage and citation validity transfer more robustly than answer extraction and refusal calibration under domain and task shifts. Overall, the findings suggest that trustworthy RAG systems should be designed as integrated retrieval--generation--validation pipelines, where candidate answers are checked against cited evidence before final delivery. Future studies should strengthen evidence utilization through improved span selection, entailment-based validation, and bounded multi-pass regeneration.

% \section*{Statements and Declarations}

% \subsection*{Declaration of AI-Assisted Tools}
% Artificial intelligence-assisted tools, including SciSpace, were used to support language editing and manuscript organization during manuscript preparation. These tools were not used to generate research data, alter experimental results, conduct statistical analysis, or replace author interpretation. All AI-assisted content was reviewed, edited, and verified by the authors, who take full responsibility for the final manuscript.

% \subsection*{Funding}
% This research did not receive any specific grant from funding agencies in the public, commercial, or not-for-profit sectors.

% \subsection*{Data Availability}
% The datasets used in this study include a controlled housing-law evaluation set constructed by the authors, PrivacyQA, and CUAD. PrivacyQA and CUAD are publicly available from their respective repositories. The controlled evaluation questions, processed metadata, and evaluation scripts may be made available upon reasonable request, subject to source-document licensing restrictions.

% \subsection*{Ethics Approval and Consent to Participate}
% This article does not contain any studies with human participants or animals performed by any of the authors. Ethical approval and informed consent were not required for this purely computational study using public and non-identifiable administrative text data.

% \subsection*{Consent to publish}  
% Not applicable.

% \subsection*{Competing interests}
% The authors declare no competing interests.

\FloatBarrier
\bibliography{references}

\end{document}